\documentclass[10pt,twocolumn,letterpaper]{article}

\usepackage[pagenumbers]{cvpr} 

\usepackage{graphicx}
\usepackage{amsmath}
\usepackage{amssymb}
\usepackage{booktabs}

\usepackage{amsthm,amsmath,amssymb}
\usepackage{color}
\usepackage{mathrsfs}
\usepackage[ruled]{algorithm2e} 
\usepackage{stfloats}
\usepackage{amsmath}
\usepackage{amssymb}
\usepackage{bm}
\usepackage{booktabs}   

\usepackage[pagebackref,breaklinks,colorlinks]{hyperref}

\def\cvprPaperID{5875} 
\def\confName{CVPR}
\def\confYear{2022}

\begin{document}

\title{Associative Adversarial Learning Based on Selective Attack}

\author{Runqi Wang
\and
Xiaoyue Duan
\and
Baochang Zhang
\and
Song Xue
\and
Wentao Zhu
\and
David Doermann
\and
Guodong Guo
}
\maketitle

\begin{abstract}
A human's attention can intuitively adapt to corrupted areas of an image by recalling a similar uncorrupted image they have previously seen. This observation motivates us to improve the attention of adversarial images by considering their clean counterparts. To accomplish this, we introduce Associative Adversarial Learning (AAL) into adversarial learning to guide a selective attack.
We formulate the intrinsic relationship between attention and attack (perturbation) as a coupling optimization problem to improve their interaction.  This leads to an attention backtracking algorithm that can effectively enhance the attention's adversarial robustness.  Our method is generic and can be used to address a variety of tasks by simply choosing different kernels for the associative attention that select other regions for a specific attack. Experimental results show that the selective attack improves the model's performance. We show that our method improves the recognition accuracy of adversarial training on ImageNet by 8.32\% compared with the baseline. It also increases object detection mAP on PascalVOC by 2.02\% and recognition accuracy of few-shot learning on miniImageNet by 1.63\%. 

\end{abstract}

\section{Introduction}\label{sec:introduction}

Adversarial attacks on image recognition systems~\cite{szegedy2013intriguing} add small perturbations to images. A slight perturbation of a benign input can drastically change the performance of the  deep neural network (DNN) with high confidence. The change indicates that the DNN models are not learning the fundamental visual concepts and lack the robustness to adversarial attacks. It is common to achieve such robustness by detecting and rejecting adversarial examples~\cite{xu2017feature, meng2017magnet, ma2018characterizing}. As a kind of proactive defense, adversarial training incorporates adversarial examples into the training process to improve the model's adversarial robustness~\cite{mustafa2019adversarial, shafahi2019adversarial}. To make the learned models more dependable, the adversarial robustness of the network is crucial~\cite{shafahi2019adversarial}.

Although traditional adversarial training strategies can improve the recognition performance of attacked images, their performance on clean images will be affected~\cite{kurakin2016adversarial,tramer2017ensemble,na2017cascade}. Furthermore, their potential for other challenging tasks such as few-shot learning and object detection has not been thoroughly investigated. 
The reason lies partially in the fact that traditional adversarial training methods indiscriminately add adversarial attacks to the whole input image without a region selection process. 
It has been illustrated in~\cite{long2015fully} that models are sensitive to global and local adversarial attacks, motivating us to investigate selective attacks on critical local regions. 
Furthermore, attacking according to the importance of image regions is a kind of input distribution shift~\cite{zhang2021delving}, and promoting the model's learning of distribution shift is an effective way to improve the model's adversarial robustness. Spatial attention~\cite{woo2018cbam} has been introduced to determine critical image regions to achieve this promotion. However, spatial attention can only capture areas directly related to the goal of the task on clean images~\cite{walle2021asso}. When the image is attacked, unimportant regions may be given too much attention, and thus the critical areas may be ignored by the attention due to the existence of perturbation. Therefore, attention needs to be improved to identify  regions for a selective attack better. This motivates us to introduce our Associative Adversarial Learning (AAL) method to enhance the  attention's selection of critical image regions and  achieve better adversarial robustness across different tasks.

\begin{figure}
    \centering
    \includegraphics[width=0.9\linewidth]{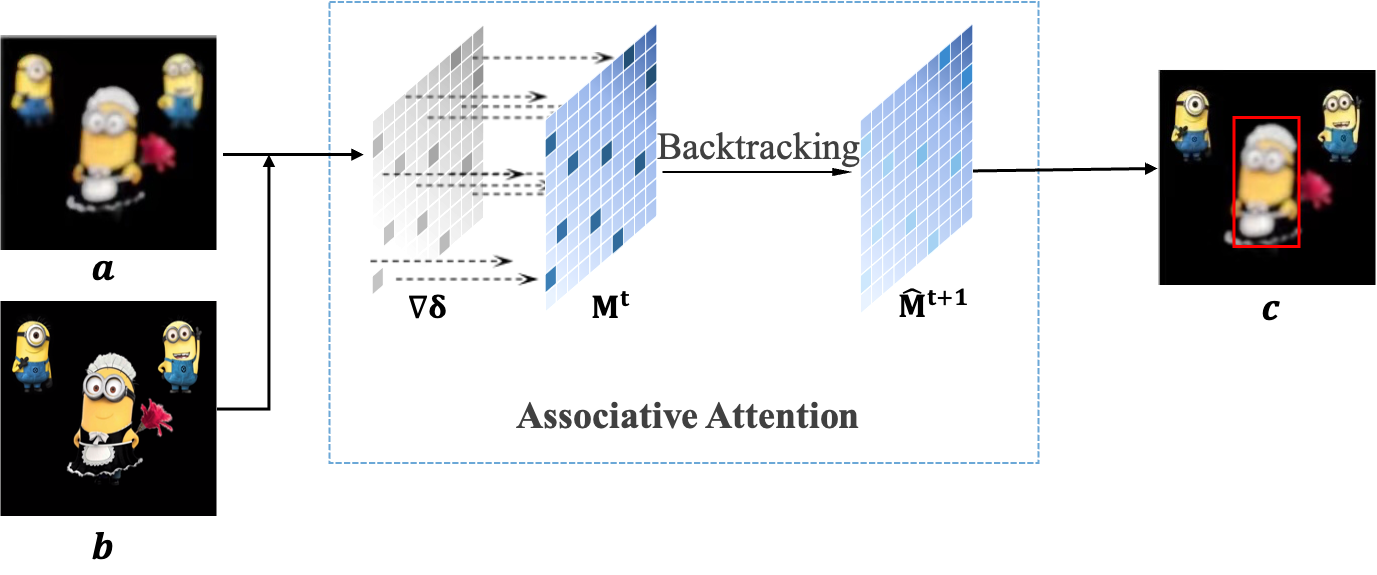}
    \caption{Associative Adversarial Learning uses an associative attention mechanism to associate adversarial images with clean images to select critical image regions to attack. 
    Figure $\bm{a}$ shows the adversarial image, $\bm{b}$ shows the clean counterpart, and $\bm{c}$ shows the image after the selective attack where the more salient area (red bounding box) is given more attack.}
    \label{fig:moti}
\end{figure}

Human-like knowledge can be realized using the same associative processes previously established for learning similar, more specific knowledge features ~\cite{behrens2008associative, shanks1995psychology, thompson1997associative}. Intuitively, when humans see a perturbed image area, they can recall a related clean image and correct the region affected by the perturbation. Inspired by this, we introduce associative attention into adversarial training, which associates the adversarial image with its clean counterpart to improve the representation of deep features and select critical regions to attack, 
as shown in Fig.~\ref{fig:moti}. We formulate the intrinsic relationship between attention and perturbation (attack) as a coupling optimization problem to improve their interaction~\cite{heide2015fast,yang2017image}. In particular, the perturbation for a traditional adversarial attack is used for attention backtracking to enhance the attention's adversarial robustness, while the improved attention is used to locate critical regions for the selective attack. This leads to a mechanism that utilizes associative attention to select critical areas to apply perturbations. The strength of such an approach is that it can seamlessly handle different tasks, including adversarial training, object detection, and few-shot learning. What differs in each task is the selection of the critical regions on which the attack is carried out. For example, an attack on the image background can strengthen the feature extraction capability of the few-shot learning model, while attacking the foreground can improve the generalization and robustness of the model in adversarial training and object detection. To this end, we introduce kernelized versions of association attention for different tasks to choose background or foreground for a selective attack. Our contributions are summarized as follows:

\begin{itemize}
\item 
We describe a generic adversarial training method called \textit{Associative Adversarial Learning} (AAL) as the first attempt to associate adversarial images with clean images. 
\item
We introduce a backtracking method to efficiently learn associative attention to guide selective attacks and facilitate optimization. 
Our method is flexible and practical for different tasks, as it selectively adds perturbations on other image regions.

\item
Extensive experiments demonstrate that our method achieves better performance than prior arts. AAL improves the recognition accuracy of adversarial training on ImageNet for adversarial images by 8.32\% compared with the baseline. 
\end{itemize}

\section{Related Work}

\textbf{Adversarial training and attacks.}
The success of deep learning models \cite{bengio2017deep} has been demonstrated on various computer vision tasks, such as image recognition \cite{he2016deep}, instance segmentation \cite{long2015fully}, and object detection \cite{szegedy2015going}. However, recent research has revealed that neural networks show fragility to adversarial examples, and existing deep models have difficulty resisting adversarial attacks \cite{carlini2017towards,goodfellow2014explaining,szegedy2013intriguing}. Therefore, models deployed in real-world scenarios are vulnerable to adversarial attacks \cite{liu2016delving}. Many other attack methods have been proposed following the discovery of adversarial examples \cite{szegedy2013intriguing}. \cite{goodfellow2014explaining} proposed the Fast Gradient Sign Method (FGSM) to generate adversarial examples, and \cite{kurakin2016adversarial} put forward the Basic Iterative Method (BIM), which requires multiple smaller FGSM steps to improve FGSM. Later, \cite{madry2017towards} proposed the Projected Gradient Descent (PGD) method to execute the adversarial attack, which is a variant of BIM with uniform random noise as initialization.

Many methods have also been proposed to defend against these attacks \cite{szegedy2013intriguing,cubuk2017intriguing}. 
Some methods eliminate the adversarial perturbation of the image before inputting the image into the network \cite{liao2018defense,wang2019bilateral,YeAdversarial,athalye2018obfuscated,gupta2019ciidefence}. 
However, adversarial training \cite{kurakin2016adversarial,tramer2017ensemble,na2017cascade} is considered one of the most effective defenses. Retraining by adding adversarial samples to the training data makes the model more capable of defending against attacks. These methods, however, typically add global perturbations to image samples.
Unfortunately, global attacks in adversarial training affect critical areas of the image and the data distribution~\cite{nakka2020indirect}. 


\textbf{Attention.}
Attention can be interpreted as a means of biasing the allocation of available computational resources towards the most informative components of a signal~\cite{olshausen1993neurobiological,itti1998model,itti2001computational,larochelle2010learning,mnih2014recurrent,vaswani2017attention}. Attention mechanisms have demonstrated their utility across many tasks, including localization and understanding in images~\cite{cao2015look,jaderberg2015spatial}, image segmentation~\cite{rassadin2020deep}, image captioning~\cite{xu2015show,chen2017sca}, and lip reading~\cite{son2017lip}, for example. These applications can be viewed as operators following one or more layers representing higher-level abstractions for adaptation between modalities. Some works provide interesting studies into the combined use of spatial and channel attention~\cite{wang2017residual,woo2018cbam}. For example, based on hourglass modules~\cite{newell2016stacked}, \cite{wang2017residual} introduces a powerful trunk-and-mask attention mechanism that is inserted between the intermediate stages of deep residual networks.


This paper introduces a generic adversarial learning method based on selective attacks and associative attention in the same framework. Our approach can be effectively adapted to various tasks, including adversarial training, object detection, and few-shot learning, by setting up different critical regions for attack based on specifically designed kernel functions. Experiment results show that our Associative Adversarial Learning significantly improves system performance on these tasks. 

\begin{figure*}[t]
    \centering
    \includegraphics[width=0.7\textwidth]{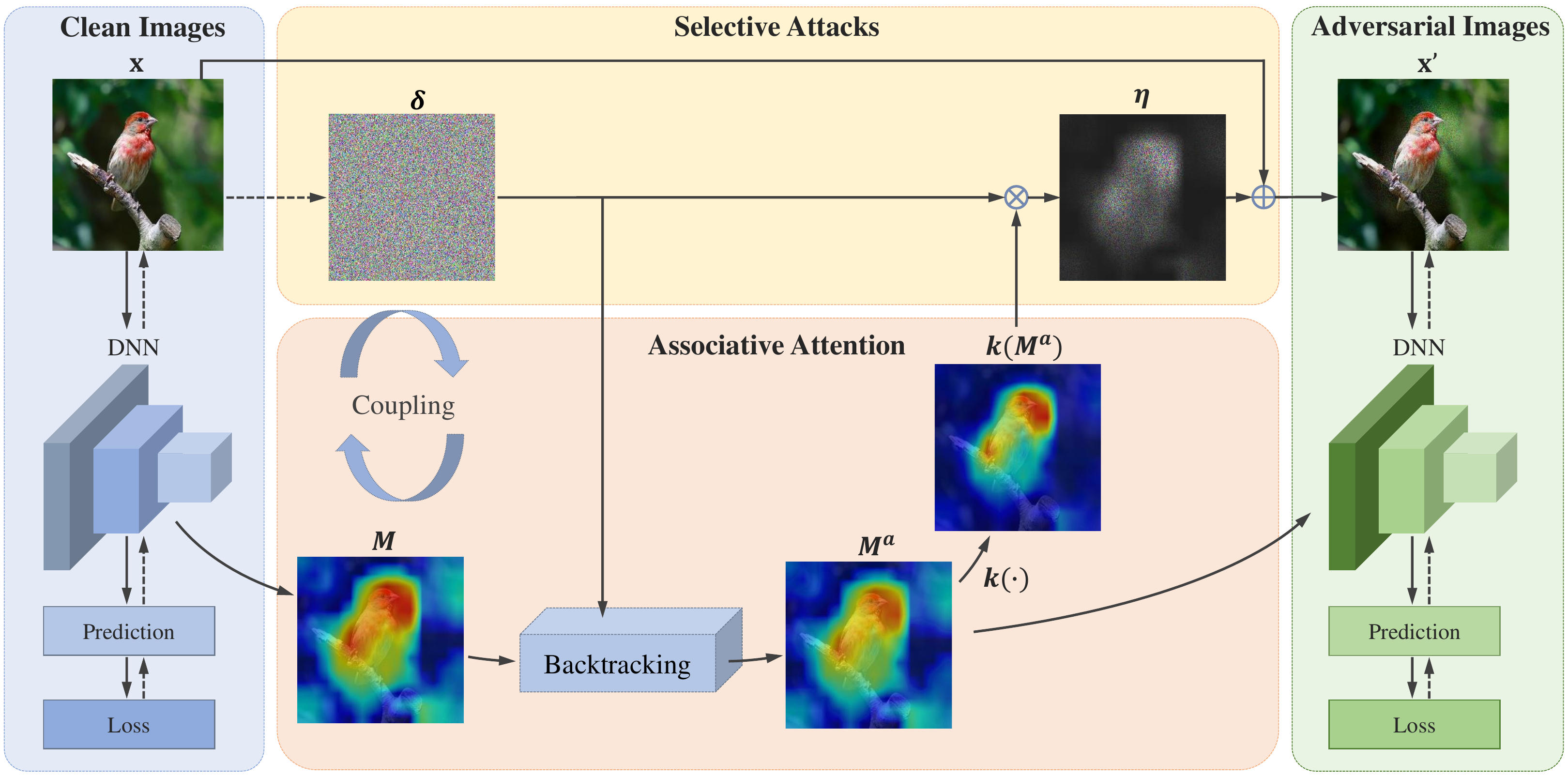}
    \caption{The framework of AAL.   The coupling information between perturbation and spatial attention is leveraged to build our kernelized associative attention, guiding selective attacks to improve adversarial robustness.}
    \label{fig:attention_bi}
\end{figure*}

\section{Associative Adversarial Learning}
\label{sec:3}

In this section, we present our proposed Associative Adversarial Learning (AAL) framework 
as shown in Fig.~\ref{fig:attention_bi}.  The figure shows how associative attention guides selective attacks on critical regions by considering adversarial samples with their clean counterparts and thus improving the model's adversarial robustness. We elaborate on the AAL method below.

\subsection{Selective Attacks}\label{sec3.2}


A significant effect on instance segmentation tasks has been observed by attacking critical areas of the input image~\cite{nakka2020indirect}. Inspired by this, we explore a selective attack approach for different visual tasks and find that attacking local data enhances performance. For example, an attack on the image background can weaken its effect on model predictions and strengthen the feature extraction capability of the model. Alternatively, attacking the foreground prevents overfitting and improves the generalization and robustness of the model. Our method can capture areas reflecting the crucial and correct content in adversarial images and selectively attack these areas. 


We start with a discussion of adversarial samples. The typical input in adversarial training is $\mathbf{x'}=\mathbf{x}+\bm{\delta}$, where $\mathbf{x}\in\mathbb{R}^{H\times H}$(the shape of inputs is resized to $H\times H$) is the clean counterpart of the adversarial input $\mathbf{x'}$, and $\bm{\delta}$ is the perturbation added to the clean input. Traditional adversarial training methods typically add $\bm{\delta}$ to the whole image indiscriminately, such as in~\cite{goodfellow2014explaining} where $\bm{\delta}$ is defined as:
\begin{equation}
    \bm{\delta} = \bm{\epsilon}\cdot \mathbf{sign}(-\nabla_\mathbf{x} G(\mathbf{W}, \mathbf{x})).
    \label{eq:delta}
\end{equation}
 Here, $\mathbf{W}$ denotes the weights of the deep model, $G$ is the cross-entropy loss function, and $\bm{\epsilon}$ is a perturbation matrix in which each element $\epsilon_{i,j}$ satisfies $\epsilon_{i,j}\in[-\epsilon, +\epsilon]$ with $\epsilon$ as the perturbation bound. With the  bound, the recognition result can be changed by adding an imperceptibly small vector whose sign is equal to the negative gradient of the input. The perturbation $\bm{\delta}$ is global and fails to select crucial areas for a local  attack which has been proved to be effective in prior arts~\cite{nakka2020indirect}. One natural choice is spatial attention~\cite{woo2018cbam}, by which  critical regions can be found for a specific attack. However, conventional attention methods fail to consider the coupling relationship between attention and perturbation for optimization, and thus their performance suffers in adversarial learning. To address the issue, we introduce kernelized associative attention to improve $\bm{\delta}$ for a selective attack as: 
\begin{equation}
    \bm{\eta} = k(\mathbf{M}^a)\circ\bm{\delta},
\label{eq:at}
\end{equation}
where $\bm{\eta}$ is the selective perturbation, $\mathbf{M}^a\in\mathbb{R}^{H\times H}$ is the associative attention,  which will be elaborated on in the next section, and $\circ$ denotes the Hadamard production. The kernel function $k(\cdot)$ is specially designed for different tasks, e.g., $k(\mathbf{M}^a)$= $\mathbf{M}^a, (\mathbf{M}^a\circ\mathbf{M}^a) $ and $(1-\mathbf{M}^a\circ\mathbf{M}^a)$.  The first two kernel functions are prone to choosing the foreground as critical regions, while the last focuses on the background.  Based on Eqs.~\ref{eq:delta} and ~\ref{eq:at}, the selectively attacked adversarial sample is obtained by:
\begin{equation}
    \begin{aligned}
    \mathbf{x'} &= \mathbf{x} + \bm{\eta} \\
    &= \mathbf{x} + k(\mathbf{M}^a)\circ (\bm{\epsilon}\cdot \mathbf{sign}(-\nabla_\mathbf{x} G(\mathbf{W}, \mathbf{x}))).\\
    \end{aligned}
    \label{eq:attacks}
\end{equation}

\begin{table*}[hb]
\centering
\setlength\tabcolsep{4pt}
\caption{Notation of attention}
\small
\begin{tabular}{cc}
\hline
$\mathbf{M}$                        & Spatial attention proposed from~\cite{woo2018cbam}, $\mathbf{M} \in \mathbb{R}^{H \times H}$                                   \\
${\hat{\mathbf{M}}}$ & Spatial attention after gradient descent \\
$\mathbf{M}^a$                        & Associative attention, proposed by ${\hat{\mathbf{M}}}$ after backtracking      \\         \hline                 
\end{tabular}
\label{tab:nota}
\end{table*}

\subsection{Associative Attention}\label{sec3.3}

In this section, we  elaborate on how associative attention is calculated. We first introduce the coupling relationship between attention and  perturbation, which misleads the learning of critical regions. We then describe an attention backtracking method to obtain robust associative attention consistent with the human association mechanism.

\subsubsection{Coupling}\label{sec3.2.1}
In Eq.~\ref{eq:at},  the attention and the perturbation are conventionally assumed to be  independent of each other. However, they are coupled. For example, the attention can  identify critical areas with a clean input image, but when the input is perturbed, the attention may be incorrectly directed to the more rewarding areas for a wrong category if the model prediction is wrong.
The introduction noted that humans use their association mechanism to modify attention by recalling clean images to promote their understanding of corrupted images. Similarly, we attempt to optimize the attention  using the coupling information between the attention and the perturbation to guide selective attacks on  critical areas and improve the model's adversarial robustness.

\subsubsection{Optimization via Backtracking}\label{sec3.2.2}

To obtain the associative attention, we first define three different attention variables $\mathbf{M}$, ${\hat{\mathbf{M}}}$ and $\mathbf{M}^a$  as shown in Tab.~\ref{tab:nota}. $\mathbf{M}$ is the original spatial attention obtained based on  clean samples according to~\cite{woo2018cbam}. When using traditional adversarial methods such as~\cite{goodfellow2014explaining} to calculate input gradients for perturbation, $\mathbf{M}$ is simultaneously updated via gradient descent into  ${\hat{\mathbf{M}}}$. ${\hat{\mathbf{M}}}$ is further backtracked by considering its coupling relationship with the  perturbation  
to achieve the final associative attention $\mathbf{M}^a$.

Based on these definitions, we define the objective function as:
\begin{equation}
\label{eq:bilinear}
\mathop {\arg \min }\limits_{\mathbf{M}} G(\Phi_{\{\mathbf{W},\mathbf{M},\bm{\delta}\}}(\mathbf{x}),y),
\end{equation}
where $\Phi_{\{ \mathbf{W},\mathbf{M},\bm{\delta}\}}$ is the DNN model with three sets of parameters $\{\mathbf{W},\mathbf{M},\bm{\delta}\}$, $G$ is the cross entropy loss function, and $\mathbf{x}$ and $y$ represent the input image and its label respectively.  
Traditional adversarial learning methods solve the appropriate perturbation and attention independently by minimizing the objective function. We instead consider the coupling information between $\mathbf{M}$ and $\bm{\delta}$ to calculate $\mathbf{M}^a$ based on gradient descent and define a new update strategy of attention based on the chain rule~\cite{petersen2008matrix} as:

	\begin{equation}
	\begin{aligned}
	\label{coupled1}
	\mathbf{M}^a 
	&= {\mathbf{M}} - \xi_1\frac{\partial G(\mathbf{M})}{\partial {\mathbf{M}}} - \xi_2 \frac{\partial G( \bm{\delta})}{\partial{\mathbf{M}} } \\
	&= {\mathbf{M}} - \xi_1\frac{\partial G(\mathbf{M})}{\partial {\mathbf{M}}} - \xi_2 Tr((\frac{\partial {G(\bm{\delta})}}{\partial \bm{\delta}})^T \frac{\partial {\bm{\delta}}}{\partial {\mathbf{M}}}), \\
	\end{aligned}
	\end{equation}
where $\xi_1$ and $\xi_2$ denote the learning rates, and $Tr(\cdot)$ represents the trace of the matrix. We further define:
\begin{equation}
    \hat{G}= (\frac{\partial G( \bm{\delta})}{\partial \bm{\delta}})^T /{\mathbf{M}},
\end{equation}
where $\hat{G}$ is defined by considering the coupling optimization problem~\cite{DBLP:cvpr/Zhuocogd}  as in Eq.~\ref{eq:bilinear}. Then we have:

    \begin{equation}
		\begin{aligned}
		\frac{\partial G( \bm{\delta})}{\partial{\mathbf{M}} }=Tr[{\mathbf{M}}  \hat{G} \frac{\partial \bm{\delta}}{\partial {\mathbf{M}}}].
		\end{aligned}
		\label{eq:6}
	\end{equation}

We denote $\hat{G}=[\hat{g}_1,...,\hat{g}_H]$, $\bm{\delta}=[\bm{\delta}_1,...,\bm{\delta}_H]$, $\mathbf{M}=[\mathbf{M}_1,...,\mathbf{M}_H]^T$.
Assuming that $\delta_{i, j}$ and $\mathbf{M}_{m, n}$ are independent when they are in different columns, we have:

	\begin{equation}
		\frac{\partial \bm{\delta}_m}{\partial \mathbf{M}}=
		\begin{bmatrix} 
		0 &...&\frac{\partial \bm{\delta}_{m}}{\partial \mathbf{M}_{1,m}}&...& 0 \\
		.&&.&&.\\
		.&&.&&.\\
		.&&.&&.\\ 
		0 &...&\frac{\partial \bm{\delta}_{m}}{\partial \mathbf{M}_{H,m}}&...& 0
		\end{bmatrix},
		\label{eq:8}
	\end{equation} 
where $\{\cdot\}_{m,n}$ denotes the element at row $m$ column $n$, and
	\begin{equation}
		\mathbf{M}  \hat{G}=
		\begin{bmatrix} 
		\mathbf{M}_1  \hat{g}_1 &...&\mathbf{M}_1  \hat{g}_n&...& \mathbf{M}_1  \hat{g}_H \\
		.&&.&&.\\
		.&&.&&.\\
		.&&.&&.\\ 
		\mathbf{M}_H  \hat{g}_1 &...&\mathbf{M}_H  \hat{g}_n&...& \mathbf{M}_H  \hat{g}_H
		\end{bmatrix}.
		\label{eq:9}
	\end{equation}
Combining Eq.~\ref{eq:8} and Eq.~\ref{eq:9}, we then have:
	\begin{equation}
		\mathbf{M}  \hat{G}\frac{\partial \bm{\delta}_m}{\partial \mathbf{M}}=
		\begin{bmatrix} 
		0 &...&\mathbf{M}_1  \sum_{n=1}^H\hat{g}_n\frac{\partial \bm{\delta}_{m}}{\partial \mathbf{M}_{n,m}}&...& 0 \\
		.&&.&&.\\
		.&&.&&.\\
		.&&.&&.\\ 
		0 &...&\mathbf{M}_H  \sum_{n=1}^H\hat{g}_n\frac{\partial \bm{\delta}_{m}}{\partial \mathbf{M}_{n,m}}&...& 0
		\end{bmatrix}.
		\label{eq:10}
	\end{equation}
Therefore, the trace of Eq.~\ref{eq:6} is then calculated by:
	\begin{equation}
		Tr[\mathbf{M}  \hat{G} \frac{\partial \bm{\delta}_m}{\partial \mathbf{M}}]=\mathbf{M}_m  \sum_{n=1}^H\hat{g}_n\frac{\partial \bm{\delta}_{m}}{\partial \mathbf{M}_{n,m}}.
		\label{eq:11}
	\end{equation}

We then define $\hat{\mathbf{M}}$ as:
\begin{equation}
    \hat{\mathbf{M}} = \mathbf{M} - \xi_1 \frac{\partial G(\mathbf{M})}{\partial \mathbf{M}}.
    \label{eq:m^t+1}
\end{equation}

Combining Eq.~\ref{coupled1}, Eq.~\ref{eq:11} and Eq.~\ref{eq:m^t+1}, 
we achieve a backtracking method for gradient update to solve $\mathbf{M}^{a}$ as:
	\begin{equation}
		\label{eq:12}
		\begin{aligned}
		\mathbf{M}^a&=\hat{\mathbf{M}}-\xi_2
		\begin{bmatrix} 
		\sum_{n=1}^H\hat{g}_n\frac{\partial \bm{\delta}_{1}}{\partial \mathbf{M}_{n,1}}\\
		.\\
		.\\
		.\\ 
		\sum_{n=1}^H\hat{g}_n\frac{\partial \bm{\delta}_H}{\partial \mathbf{M}_{n,H}}
		\end{bmatrix}    
		\circ
		\begin{bmatrix} 
		\mathbf{M}_1\\
		.\\
		.\\
		.\\ 
		\mathbf{M}_H
		\end{bmatrix}\\    
		&=\hat{\mathbf{M}}-\xi_2
		\begin{bmatrix} 
		<\hat{G},\frac{\partial \bm{\delta}_1}{\partial \mathbf{M}_1}>\\
		.\\
		.\\
		.\\ 
		<\hat{G},\frac{\partial \bm{\delta}_{H}}{\partial \mathbf{M}_H}>
		\end{bmatrix}    
		\circ
		\begin{bmatrix} 
		\mathbf{M}_1\\
		.\\
		.\\
		.\\ 
		\mathbf{M}_H
		\end{bmatrix}\\    
		&=\hat{\mathbf{M}}-\xi_2 {\bm{\gamma}} \circ \mathbf{M} \\
		&=P(\hat{\mathbf{M}},\mathbf{M}),	
		\end{aligned}
	\end{equation}
where $\xi_2$ denotes the learning rate of backtracking and $\xi_2 \bm{\gamma}$ denotes the step size of backtracking. $\bm{\gamma}$ is the linear kernel function $\hat{G} \circ \frac{\partial \bm{\delta}_{m}}{\partial \mathbf{M}_m}$, which implies the coupling relationship between perturbation and attention. To simplify the calculation, $\frac{\partial \bm{\delta}}{\partial \mathbf{M}}$ can be approximated by $\frac{\Delta \bm{\delta}}{\Delta \mathbf{M}}$, which denotes the relative change of $\bm{\delta}$ with respect to $\mathbf{M}$. Eq.~\ref{eq:12} shows that our method is actually based on a projection function to solve the coupling optimization problem by $\bm{\gamma}$. 

In this method, we consider the coupling information in $\mathbf{M}$ to backtrack $\hat{\mathbf{M}}$ and finally obtain the associative attention $\mathbf{M}^a$.
To better associate clean images and adversarial images, $\mathbf{M}^a$ needs compensation  for the attention on the severely perturbed areas in the updating process. To this end, we introduce the triggering condition for backtracking defined as: 

\begin{equation}
	\mathbf{M}^a_{m,n}=
	\begin{cases}
	P(\hat{\mathbf{M}}_{m,n},\mathbf{M}_{m,n})& if\ R(\frac{\partial {G}}{\partial \delta_{m, n}})>\zeta,\\
	\hat{\mathbf{M}}_{m,n}& otherwise,
	\end{cases}
	\label{eq:proj}
\end{equation}
where $\delta_{m, n}$ denotes the element of the perturbation matrix, $P(\cdot)$ denotes the projection function as shown in Eq.~\ref{eq:12}.  $R(\frac{\partial {G}}{\partial \delta_{m, n}})$ denotes  the ranking of $\frac{\partial {G}}{\partial \delta_{m, n}}$, which is larger than  the threshold $\zeta$ for a backtracking. 

In summary,  our associative attention is solved with a new coupling optimization method and achieves robustness to adversarial attacks based on a mechanism consistent with the human association. Notably, 
our kernelized associative attention $k(\mathbf{M}^a)$  can easily adapt  to different applications as validated in our experimental results below. Our algorithm is summarized in Alg.~\ref{alg:st}.


\begin{algorithm}
	\caption{Associative Adversarial Learning
	\label{alg:st}}
	\textbf{Input:} Training data and validation data;\\ \textbf{Hyper-parameters:} Training epoch $S$, $\xi_1=0.1$, $\xi_2=0.1$ and $t=0$; \\
	Initialize the model weights $\mathbf{W}$; \\
	\textbf{Output:} The network model; \\
	$\#$ Training an architecture for $S$ epochs: \\
	\While{$(t\leq S)$}
        {$\#$ First inference:\\
        According to~\cite{woo2018cbam}, calculate the spatial attention $\mathbf{M}$; \\
        $\#$ Back propagation:\\
        According to Eq.~\ref{eq:m^t+1}, calculate $\hat{\mathbf{M}}$;\\
        According to Eqs.~\ref{eq:12} and~\ref{eq:proj}, calculate $\mathbf{M}^a$;\\
        $\#$ Second inference:\\
        Add perturbation to the data via Eq.~\ref{eq:attacks};\\
        According to Eq.~\ref{eq:bilinear}, calculate the objective function;\\
        $\#$ Back propagation:\\
        Update weights $\mathbf{W}$;\\
        $t \leftarrow t + 1$.}
\end{algorithm}

\section{Experiments}
\label{sec:4}

To validate our AAL's effectiveness, we test it on three tasks: adversarial training, object detection, and few-shot learning. For adversarial training and object detection,
we add perturbation to the image foreground based on the kernel functions $k(\mathbf{M}^a)=\mathbf{M}^a$ and $(\mathbf{M}^a\circ\mathbf{M}^a) $.
For few-shot learning, perturbation is added to the background using $k(\mathbf{M}^a)=(1-\mathbf{M}^a\circ\mathbf{M}^a)$. Additional results are included in the supplementary files.

\subsection{Adversarial Training}
\textbf{Datasets}. 
CIFAR10 \cite{krizhevsky2009learning} is a popular dataset for image recognition, which contains 60K images, with 50K images as the training set and the remaining 10K images as the testing set. The images include ten different categories and have a resolution of $32\times32$. ImageNet ILSVRC 2012~\cite{krizhevsky2012imagenet} is a commonly used large-scale image recognition dataset that consists of 1,000 categories with 1.2M training images and 50K validation images.

\textbf{Experimental setting}. We conduct experiments on both the CIFAR10 \cite{krizhevsky2009learning}
and the ImageNet \cite{krizhevsky2012imagenet} datasets. We adopt ResNet-18~\cite{he2016deep} or WRN-34-10~\cite{zagoruyko2016wide} for CIFAR10,
and ResNet-50~\cite{he2016deep} for ImageNet.
We use the FGSM adversarial training method~\cite{goodfellow2014explaining} with the perturbation bound $\epsilon=\frac{8}{255}$ and a step size of $\frac{2}{255}$ on CIFAR10 and ImageNet, and use PGD~\cite{madry2017towards} with $\epsilon=\frac{8}{255}$, step size of $\frac{2}{255}$ and $10$ iterative steps on CIFAR10. We adopt an SGD optimizer with a momentum of 0.9 and a weight decay of $5\times 10^{-4}$. The initial learning rate $\xi_1$ is set to 0.1 for CIFAR10 and 0.2 for ImageNet (annealed down to 0 following a cosine schedule without restart). $\xi_2=0.1$ and $\zeta=0.1$. $k(\mathbf{M}^a)$, as shown in Eq.~\ref{eq:attacks}, is chosen as $({\mathbf{M}^a}\circ{\mathbf{M}^a})$, which promotes selective attacks on critical areas of the foreground areas. We train 250 epochs on CIFAR10 and 250 epochs on ImageNet. We evaluate our method under the clean dataset, the FGSM attack, and the PGD-$10$ attack.

\textbf{Results.} As is shown in Tabs.~\ref{tab:adv_cifar} 
and \ref{tab:adv_image}, AAL outperforms prior arts, including LS~\cite{szegedy2016rethinking}, FGSMR~\cite{yisen2020}, and IAAT~\cite{balaji2019instance}.

\begin{table}[t]
\centering
\setlength\tabcolsep{4pt}
\caption{Comparison of different adversarial training results on CIFAR10.}
\begin{tabular}{ccccc}
\toprule
\textbf{\begin{tabular}[c]{@{}c@{}}Training\\ method\end{tabular}} & \textbf{Backbone} & \textbf{Clean} & \textbf{FGSM} & \textbf{PGD}\\ \toprule
Clean                                                              & ResNet-18         & 95.46        &    20.18    & -            \\

FGSM       & ResNet-18         & 92.55     &  82.40   &  41.64            \\

PGD     & WRN-34-10         & 86.88  & 62.68      &47.69        \\

LS0.8         & WRN-34-10           &   87.28       &  66.09        &53.49            \\

LS0.9     & WRN-34-10         & 87.64    & 65.96         &52.82               \\

 FGSMR   &ResNet-18 &  77.10   &  -   & 60.60      \\   
 

\hline
AAL+FGSM                                                       & ResNet-18         & 95.40                                                               & 83.85     &42.89

\\
AAL+PGD                                                       & WRN-34-10         & 89.70        & 68.78   & 60.88
\\ \toprule
\end{tabular}
\label{tab:adv_cifar}
\end{table}

\begin{table}[t]
\centering
\caption{Comparison of different adversarial training results on ImageNet.}
\begin{tabular}{ccccc}
\toprule
\textbf{\begin{tabular}[c]{@{}c@{}}Training\\ method\end{tabular}} & \textbf{Backbone} & \textbf{Clean} & \textbf{FGSM}
 \\ \toprule
Clean   & ResNet-50         & 73.80    & 7.56              
\\

FGSM       & ResNet-50         &58.74   &  52.64         
\\


IAAT       & ResNet-50     & 62.71     & 56.61         
\\               

 \hline
AAL+FGSM     & ResNet-50         &  72.73 &60.96          

\\ \toprule
\end{tabular}
\label{tab:adv_image}
\end{table}

Tab.~\ref{tab:adv_cifar} compares results of AAL and other different adversarial training results on CIFAR10. Comparing AAL+FGSM training with clean training, it is worth noting that although the validation accuracy of clean training is slightly better than AAL+FGSM ($95.46\%$ vs. $95.40\%$) on the clean dataset, AAL+FGSM significantly improves the adversarial robustness of the model ($20.18\%$ vs. $83.85\%$). Moreover, after integrating AAL into FGSM (AAL+FGSM), the validation accuracy of the FGSM baseline increases by 2.85\% (from $92.55\%$ to $95.40\%$), 1.45\% (from $82.40\%$ to $83.85\%$), and 1.25\% (from $41.64\%$ to $42.89\%$) on the clean dataset, the FGSM adversarial dataset and the PGD adversarial dataset, respectively. Similarly, compared with the PGD baseline, the accuracy of AAL+PGD training increases by 2.82\% (from $86.88\%$ to $89.70\%$), 6.1\% (from $62.68\%$ to $68.78\%$) and 13.19\% (from $47.69\%$ to $60.88\%$) on the three validation datasets, respectively.




Tab.~\ref{tab:adv_image} shows the comparison results on ImageNet. AAL increases the validation accuracy of the FGSM method by $13.99\%$ (from $58.74\%$ to $72.73\%$) and $8.32\%$ (from $52.64\%$ to $60.96\%$) on the clean dataset and the FGSM adversarial dataset, respectively, with the validation accuracy ($72.73\%$) almost equal to that of clean training ($73.80\%$) on the clean dataset. 
The results indicate that AAL can indeed enhance the robustness of the model against perturbation without losing its generalization ability to recall information from clean images to promote performance effectively.

\subsection{Object Detection}
\textbf{Datasets}. The Pascal VOC~\cite{everingham2015pascal} is a collection of datasets for object detection, which consists of four categories (Vehicle, Household, Animal, and Person) and 20 small classes. MS COCO 2014~\cite{lin2014microsoft} includes 80 object categories. In our experiments, models are trained with a combination of 80k images from COCO train2014 and 35k images sampled from COCO val2014, $i.e.$, COCO trainval35k. We evaluate the remaining 5k images from COCO minival. Following the standard evaluation metric of object detection, we report the average precision (AP) for $IoU \in [0.5 : 0.05 : 0.95]$ denoted as mAP@[0.5, 0.95] and $IoU = 0.5$ denoted as mAP@0.5. We use mAP@0.5 for the Pascal VOC and mAP@[0.5, 0.95] for COCO.

\textbf{Experimental setting.} Our experiments use the SSD-512~\cite{liu2016ssd} detection architecture with ResNet-34~\cite{he2016deep} as the backbone.  ResNet-34 is pretrained on the ILSVRC ImageNet dataset~\cite{krizhevsky2012imagenet}. We fine-tune the pretrained model for 24 epochs on the object detection datasets mentioned above using SGD with a momentum of 0.9 and a weight decay of $5\times 10^{-4}$. We set the initial learning rate to $3\times 10^{-3}$ and the batch size to 32. An attack is guided to the foreground by the kernel function $({\mathbf{M}^a}\circ{\mathbf{M}^a})$ in this task. The hyper-parameters of the training of adversarial images are the same as those in the adversarial training task. The clean validation set is used for testing.

\begin{table}[t]
\centering
\caption{The results of AAL adopted in object detection based on SSD-512.}
\begin{tabular}{cccc}
\toprule
\textbf{\begin{tabular}[c]{@{}c@{}}Training\\ method\end{tabular}} & \textbf{Backbone} & \textbf{Dataset}    & \textbf{mAP}   \\\toprule
Clean                                                              & ResNet-34         & Pascal VOC & 75.30 \\
AAL+FGSM                                                                & ResNet-34         & Pascal VOC & \textbf{77.32} \\
Clean                                                              & ResNet-34         & MS COCO    &   31.60    \\
AAL+FGSM                                                              & ResNet-34         & MS COCO    & \textbf{33.10} 
\\ \toprule
\end{tabular}
\label{tab:detect}
\end{table}

\textbf{Results.} As shown in Tab.~\ref{tab:detect}, AAL is validated on both benchmark datasets for object detection. Compared with the baselines, AAL increases mAP by 2.02\% on Pascal VOC and 1.50\% on MS COCO 2014, respectively. Conclusively, through a selective attack on images guided by associative attention, AAL effectively enhances the performance on object detection tasks.

\subsection{Few-shot Learning}

Through an association mechanism, humans can learn to recognize new classes with the help of a few labeled samples of different but similar classes~\cite{fei2006one}. Few-shot learning aims to mimic this ability by combining it with deep learning to recognize unlabeled samples with few labeled instances. Therefore, we evaluate our AAL method on few-shot learning to address the effectiveness of our associative attention mechanism. \cite{finn2017model} proposes a model agnostic meta-learning (MAML) algorithm to find parameters that are sensitive to changes in the task with a small number of samples, which is a milestone in the few-shot learning development. 
In IER~\cite{rizve2021exploring}, the method of "multi-task learning" is used to integrate the loss functions of "learning Invariant" and "learning Equivariant" in self-supervised learning, and thus learns a robust feature extraction model to improve the recognition accuracy of few-shot learning tasks. Furthermore, we compare 
Proto-Net~\cite{snell2017prototypical}, 
RelationNet~\cite{sung2018learning},   
Shot-Free~\cite{ravichandran2019few}, 
MetaOptNet~\cite{lee2019meta}, 
Boosting~\cite{gidaris2019boosting}, 
Fine-tuning~\cite{dhillon2019baseline}, 
LEO-trainval~\cite{rusu2018meta},
Deep DTN~\cite{chen2020diversity}, 
AWGIM~\cite{guo2020attentive}, 
DSN-MR~\cite{simon2020adaptive} 
and 
RFS~\cite{tian2020rethinking} 
with our method AAL in Tabs.~\ref{tabmini} and \ref{tabtier}. Unlike the adversarial training task, we conduct a selective attack on the background areas of the images, and thus the kernel function $k(\mathbf{M}^a)$ in Eq.~\ref{eq:attacks} is chosen as $(1-{\mathbf{M}^a}\circ{\mathbf{M}^a})$. 

\begin{table}[t]
    \centering
    \caption{Average 5-way few-shot recognition accuracy with 95\% confidence intervals on miniImageNet dataset.}
    \begin{tabular}{cccc}
    \toprule
    \textbf{Methods} & \textbf{Backbone} & \textbf{1-shot} & \textbf{5-shot} \\ \toprule
    MAML            & 32-32-32-32       & 48.70           & 63.11            \\
    Proto-Net        & 64-64-64-65       & 49.42           & 68.20           \\
    RelationNet     & 54-96-128-256     & 50.44           & 65.32           \\
    Shot-Free     & ResNet-12         & 59.04           & 77.64           \\
    MetaOptNet      & ResNet-12         & 62.64           & 78.63           \\
    Bossting     & WRN-28-10         & 63.77           & 80.70           \\
    Fine-tuning     & WRN-28-10         & 57.73           & 78.17           \\
    LEO-trainval  & WRN-28-10         & 61.76           & 77.59           \\
    Deep DTN      & ResNet-12         & 63.45           & 77.91           \\
    AWGIM           & WRN-28-10         & 63.12           & 78.40           \\
    DSN-MR           & ResNet-12         & 64.60           & 79.51           \\
    RFS-Simple   & ResNet-12         & 62.02           & 79.64           \\
    IER             & ResNet-12         & 66.82           & 84.35           \\ \hline
  AAL    & ResNet-12         & \textbf{68.45}           & \textbf{85.02}
               \\
    \bottomrule
    \end{tabular}
    \label{tabmini}
\end{table}

\begin{table}[t]
\centering
\caption{Average 5-way few-shot recognition accuracy with 95\% confidence intervals on tieredImageNet dataset.}
\begin{tabular}{cccc}
\toprule
\textbf{Methods} & \textbf{Backbone} & \textbf{1-shot} & \textbf{5-shot} \\ \toprule
    MAML            & 32-32-32-32       & 51.67           & 70.30            \\
    Proto-Net      & 64-64-64-64       & 53.31           & 72.69           \\
    RelationNet     & 54-96-128-256     & 54.48           & 71.32           \\
    Shot-Free    & ResNet-12         & 63.52           & 82.59           \\
    MetaOptNet      & ResNet-12         & 65.99           & 81.56           \\
    Boosting     & WRN-28-10         & 70.53           & 84.98           \\
    Fine-tuning  & WRN-28-10         & 66.58           & 85.55           \\
    LEO-trainval   & WRN-28-10         & 66.33           & 81.44         \\
    AWGIM           & WRN-28-10         & 67.69           & 72.82           \\
    DSN-MR           & ResNet-12         & 67.39           & 82.85           \\
    RFS-Simple       & ResNet-12         & 62.02           & 79.64           \\
    IER             & ResNet-12         & 71.87           & 86.82           \\ \hline
    AAL & ResNet-12  &       \textbf{72.65}     & \textbf{87.33}
               \\
  
    \bottomrule
    \end{tabular}
    \label{tabtier}
\end{table}

\textbf{Datasets}. The miniImageNet \cite{krizhevsky2012imagenet} dataset is a subset of 100 classes randomly selected from the ImageNet dataset, with each class containing 600 images. Following~\cite{ravi2016optimization}, the dataset is divided into training, validation, and test sets, with 64, 16, and 20 classes, respectively. The tieredImageNet \cite{krizhevsky2012imagenet} dataset is also a large subset of Imagenet with 608 classes. Unlike miniImageNet, it has a hierarchical structure of broader categories of high-level nodes in ImageNet. This set of nodes is partitioned into 20, 6, and 8 disjoint sets of training, validation, and testing nodes, and the corresponding categories form the corresponding meta-sets, respectively. Therefore, samples from the training categories have semantic information different from those in testing categories, making the dataset more challenging and realistic for few-shot learning.

\textbf{Experimental setting}.
We use a ResNet-12~\cite{he2016deep} network as our base learner to conduct experiments on miniImageNet and tieredImageNet datasets. For both datasets, we use a SGD optimizer with a momentum of 0.9, a weight decay of $5\times 10^{-4}$, and we set the initial learning rates $\xi_1=0.1$, $\xi_2=0.1$, and $\zeta=0.1$. For experiments on miniImageNet, we train for 65 epochs, and the learning rates are decayed by a factor of 10 after the first 60 epochs. We train for 60 epochs for experiments on tieredImageNet, with the learning rates decayed by a factor of 10 for three times after the first 30 epochs.
We selectively attack the background areas of images in the support set during training, consisting of 5 images for each category in 5-shot learning or one image for each category in 1-shot learning. Then we evaluate our model on the query set.

\textbf{Results.} We present our results on two popular benchmark few-shot learning datasets in Tabs.~\ref{tabmini} and \ref{tabtier}, which show that our proposed method consistently outperforms all of the other methods in both 5-way 1-shot and 5-way 5-shot tasks and on both datasets. For example, on the miniImageNet dataset, the accuracy of our learning approach exceeds that of IER~\cite{rizve2021exploring} by 1.63\% and 0.67\% in 1-shot learning and 5-shot learning tasks, respectively. Our method also exceeds IER in both tasks on the tieredImageNet dataset. This demonstrates that by introducing associative attention, which performs selective attacks on image background, our AAL method effectively learns crucial image features with a minimal number of samples, achieving advanced performance in few-shot learning tasks. More comparisons are included in the supplementary.

\subsection{Ablation Study}

\begin{figure}
    \centering
    \includegraphics[width=1\linewidth]{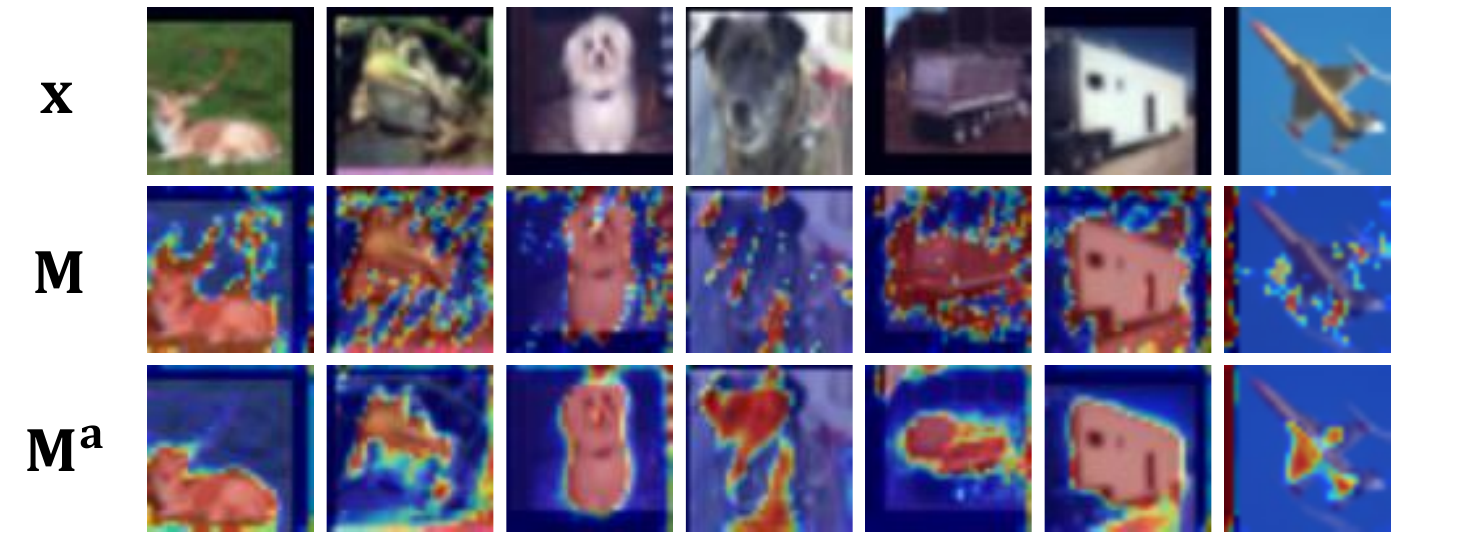}
    \caption{Visualization of the spatial attention $\mathbf{M}$ and its corresponding associative attention $\mathbf{M}^a$ on CIFAR10.}
    \label{fig:visual_atten}
\end{figure}

\textbf{Associative attention.} We compare our associative attention with the classical attention mechanism,
CBAM~\cite{woo2018cbam}. 
These two types of attention are added after the first convolutional layer of the ResNet-50 backbone~\cite{he2016deep}. 
CBAM adopts both clean training and FGSM adversarial training methods, while the associative attention model adopts FGSM adversarial training only. We use ImageNet100~\cite{rajasegaran2020itaml} as the dataset, consisting of 100 categories randomly selected from ImageNet, and each category in ImageNet100 has 500 training images and 50 validation images. Models of these two types of attention are evaluated on the clean set and the FGSM adversarial set of ImageNet100. The results are demonstrated in Tab.~\ref{tab:attention}, which shows that our associative attention increases the recognition accuracy by 3.28\% and 15.29\%, respectively on the clean dataset and the FGSM adversarial dataset of ImageNet100 compared with the FGSM baseline adopting CBAM, 
thus proving the adversarial robustness and the effectiveness of the associative attention. We further visualize CBAM and our associative attention in Fig.~\ref{fig:visual_atten}. The results of visualization illustrate that compared with spatial attention, which is more easily interfered by perturbation, our associative attention does focus on critical image regions with discriminative object features more robustly.

\begin{table}[]
\centering
\caption{Comparison of our associative attention and CBAM on ImageNet100.}
\begin{tabular}{cccc}
\toprule
\textbf{\begin{tabular}[c]{@{}c@{}}Attacking\\ method\end{tabular}} & \textbf{Attention} & \textbf{Clean} & \textbf{FGSM}  \\ \toprule
Clean                                                              & CBAM         & 89.12                                                               & 23.16                                                                     \\


FGSM                                                      & CBAM         & 86.62                                                               & 73.90                                                                     \\
FGSM                                                       & Associative Attention         & \textbf{89.90}                                                                & \textbf{89.19}       

\\ \toprule
\end{tabular}
\label{tab:attention}
\end{table}

\textbf{The kernel function $\bm{k(\cdot)}$.} We test three kernel functions mentioned in Section~\ref{sec3.2} in adversarial training. According to the results in  Tab.~\ref{tab:local}, the validation accuracy of $(1-{\mathbf{M}^a}\circ{\mathbf{M}^a})$ is much lower than that of $\mathbf{M}^a$ and $({\mathbf{M}^a}\circ{\mathbf{M}^a})$, with $({\mathbf{M}^a}\circ{\mathbf{M}^a})$ the best. The results prove that 
only attacking the most critical region in the foreground can effectively improve the model's generalization and adversarial robustness.

\begin{table}[]
\centering
    \caption{Ablation study about $k(\mathbf{M}^a)$ on ImageNet100.}
\begin{tabular}{cccc}
\toprule
\textbf{\begin{tabular}[c]{@{}c@{}}Training
\\method\end{tabular}} & \textbf{\begin{tabular}[c]{@{}c@{}}Kernel\\ functions\end{tabular}} & \textbf{Clean} & \textbf{FGSM} \\ \toprule
AAL                                                       & $\mathbf{M}^a$                                               & 89.16                                                               & 87.28                                                                     \\
AAL                                                       & $1-{\mathbf{M}^a}\circ{\mathbf{M}^a}$                     & 89.20                                                               & 51.12                                                                     \\

AAL                                                       & ${\mathbf{M}^a}\circ{\mathbf{M}^a}$                      & \textbf{89.90}                                                                & \textbf{89.19}\\                                                             \toprule          
\end{tabular}
\label{tab:local}
\end{table}

\begin{figure}
    \centering
    \includegraphics[width=1\linewidth]{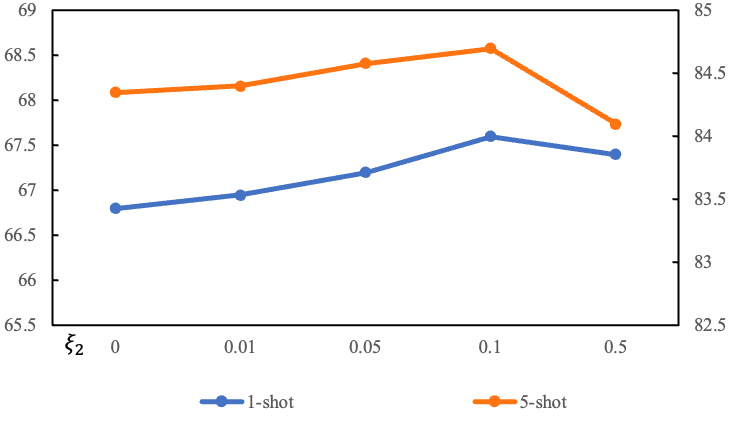}
    \caption{Associative attention vs spatial attention ($\xi_2=0$). The results of few-shot learning on miniImageNet show that the system performance is improved by backtracking, especially when $\xi_2=0.1$.}
    \label{fig:x_2}
\end{figure}

\textbf{Backtracking optimization.} To verify the impact of optimization via backtracking, we perform an ablation study in the few-shot learning task on miniImageNet by varying the value of the backtracking learning rate $\xi_2$. We present the results in Fig.~\ref{fig:x_2}. Compared with no backtracking ($\xi_2=0$), introducing the backtracking process to optimize the attention can lead to better performance of AAL on both 1-shot and 5-shot learning tasks, proving the effectiveness of backtracking optimization. Besides, the performance initially improves as $\xi_2$ increases and reaches its saturation at $\xi_2=0.1$, indicating that backtracking with a suitable learning rate is crucial for better performance. Based on the results, we set the value of $\xi_2$ to 0.1 in our experiments.

\section{Conclusion}

In this paper, we propose Associative Adversarial Learning (AAL) to solve various  visual tasks. To conduct selective attacks on critical image regions, we lead associative attention, which is solved based on a new parameter optimization method,  backtracking,  by considering the coupling relationship between perturbation and attention from a new viewpoint. Our methods achieve state-of-the-art results on adversarial training and few-shot learning, and  improve the performance in the detection task.  In future work, we will explore the potential of our AAL method on more applications.\footnote{The source code will be available upon publication.}

{\bf Limitations and potential negative social impact.} AAL  is extensively tested on various visual tasks in this paper and can be further tested on more applications if we have enough computational resources and have to set it as our future work.  AAL does not have a direct negative social impact since it is a defense method against attacks and improves the robustness of the model, but it should be prevented from being applied to harmful applications, such as being adopted to train or generate more aggressive and harmful perturbations.

{\small
\bibliographystyle{ieee_fullname}
\bibliography{Associative_Learning}
}

\end{document}